%% file: arxiv_submission.tex
\documentclass[10pt,twocolumn,letterpaper]{article}
\usepackage{threeparttable}
 \usepackage[pagenumbers]{cvpr} 

\definecolor{cvprblue}{rgb}{0.21,0.49,0.74}
\usepackage[pagebackref,breaklinks,colorlinks,allcolors=cvprblue]{hyperref}

\def\paperID{5690} 
\def\confName{CVPR}
\def\confYear{2025}

\title{REO-VLM: Transforming VLM to Meet Regression Challenges in Earth Observation}

\author{
Xizhe Xue$^{a,b}$, Guoting Wei$^{c}$, Hao Chen$^{b}$, Haokui Zhang$^{c}$, Feng Lin$^{c}$, Chunhua Shen$^{b*}$, Xiao Xiang Zhu$^{a, d*}$
\\
$^{a}$ Technical University of Munich, Germany~~~$^{b}$ Zhejiang University, China~~~\\
$^{c}$ Lighthouse~~~$^{d}$ Munich Center for Machine Learning
}

\begin{document}
\maketitle

\input{version3/0_abstract} 
\input{version3/1_intro}

\input{version3/2_relatedwork}
\input{version3/3_method}

\input{version3/4_experiment}

\input{version3/5_conclusion}

{
    \small
    \bibliographystyle{ieeenat_fullname}
    \bibliography{main}
}
\input{version3/appendix}
\end{document}

%% file: version3/0_abstract.tex
\begin{abstract}

The rapid evolution of Vision Language Models (VLMs) has catalyzed significant advancements in artificial intelligence, expanding research across various disciplines, including Earth Observation (EO). While VLMs have enhanced image understanding and data processing within EO, their applications have predominantly focused on image content description. 
This limited focus overlooks their potential in geographic and scientific regression tasks, which are essential for diverse EO applications. To bridge this gap, this paper introduces a novel benchmark dataset, called \textbf{REO-Instruct} to unify regression and generation tasks specifically for the EO domain. 
Comprising 1.6 million multimodal EO imagery and language pairs, this dataset is designed to support both biomass regression and image content interpretation tasks. Leveraging this dataset, we develop \textbf{REO-VLM}, a groundbreaking model that seamlessly integrates regression capabilities with traditional generative functions. By utilizing language-driven reasoning to incorporate scientific domain knowledge, REO-VLM goes beyond solely relying on EO imagery, enabling comprehensive interpretation of complex scientific attributes from EO data. This approach establishes new performance benchmarks and significantly enhances the capabilities of environmental monitoring and resource management.

\end{abstract}

%% file: version3/1_intro.tex
\section{Introduction}

Earth Observation (EO) data has become a essential resource for diverse scientific research, crucial in areas such as ecological monitoring~\cite{allenm3leo}, weather forecasting~\cite{nguyen2024climatelearn}, disaster response and population dynamics analysis~\cite{batista2020uncovering}. These researches require not only a thorough understanding of image content but also precise regression of the real-world scientific attributes they represent~\cite{benson2024multi,hamilton2024combining}. Regression, in this context, involves modeling the relationship between environmental metrics and relevant features, enabling accurate predictions and deeper insights into complex geospatial or  ecological processes.

\begin{figure}
    \centering
    \includegraphics[width=1.0\linewidth]{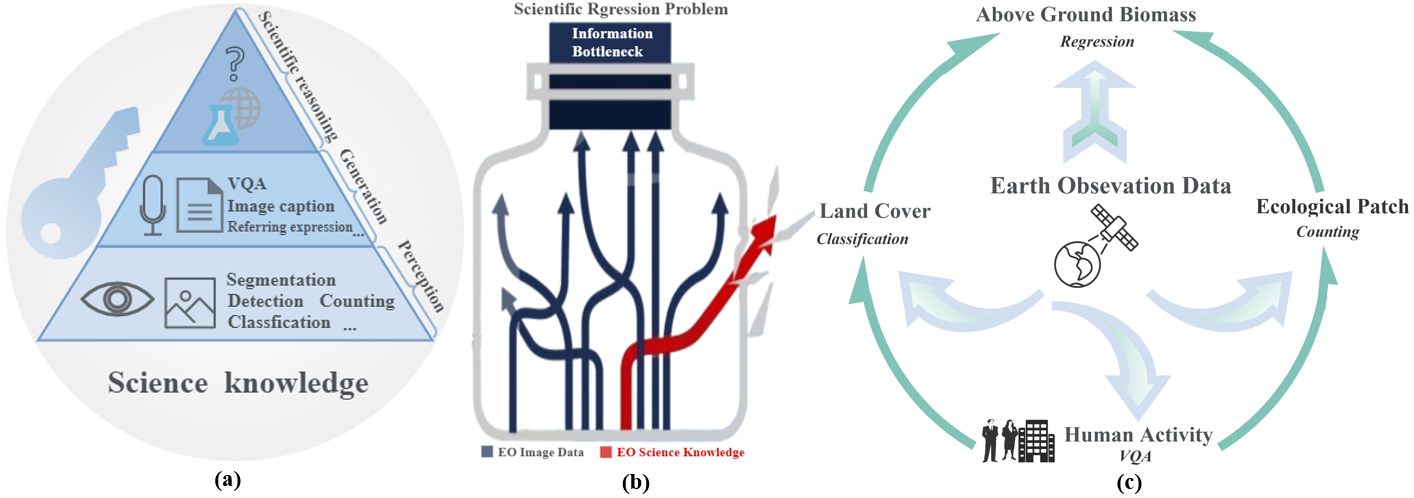}
\caption{ Motivations of VLMs for EO regression.
\textbf{(a). Hierarchical structure of VLM capabilities}: From basic perception tasks to higher-order reasoning tasks;
\textbf{(b). Advantages of VLM for EO regression tasks}: By integrating scientific domain knowledge with EO image data, VLMs overcome the information bottleneck of traditional image-only regression models, enabling deeper insights and improved scientific reasoning;
\textbf{(c). Interplay between regression and generation tasks}: Using AGB estimation as an example, the intrinsic link between regression and generation targets allows collaborative processing in a unified framework, enhancing prediction accuracy and reliability.}
    \label{fig:motivation_fig}
\end{figure}



Against this backdrop, the emergence of Vision Language Models (VLMs)~\cite{ouyang2022training,radford2021learning,liu2024visual,achiam2023gpt} offers promising directions for advancing EO analysis. As illustrated in Figure~\ref{fig:motivation_fig}(a), existing VLM applications in EO primarily focus on perception tasks (e.g., detection) and generation tasks (e.g., text-based descriptions). However, the potential of VLMs for scientific regression tasks, such as predicting environmental attributes, has received limited attention, despite its essential role in EO applications.

This work investigates the potential of VLMs for scientific regression in EO, with a focus on Above Ground Biomass (AGB) estimation~\cite{lang2023high,sialelli2024agbd}, a critical ecological indicator. As illustrated in Figure~\ref{fig:motivation_fig}(b), VLMs offer unique advantages, such as the ability to integrate multimodal EO data and incorporate domain knowledge to enhance regression accuracy. For instance, understanding the underlying land cover distribution and related characteristics improves AGB predictions. Figure~\ref{fig:motivation_fig}(c) illustrates the intrinsic relationships  between regression targets (e.g., AGB values) and generation goals (e.g., land cover classification). Despite the potential of jointly modeling regression and generation tasks to enhance prediction accuracy, fully harnessing VLMs for scientific regression requires addressing several key technical challenges:

\begin{itemize} 
\item \textbf{Inadequate data representation and training signals:} The pretraining data and objectives of VLMs are usually language-focused, not optimized for learning fine-grained numeric relationships, making it challenging to bridge the gap between rich input features and precise numeric outputs. 

\item \textbf{Symbolic output vs. continuous values:} VLMs generate discrete tokens rather than directly modeling continuous numerical values. They focus on semantic meaning while overlooking the mathematical aspects, making it challenging to achieve precise and stable regression results

\item \textbf{Error accumulation in multi-step generation:} Numeric values are often represented as multiple tokens. Any prediction error in one token can propagate through the sequence, undermining the accuracy of the final numerical result. 

\item \textbf{Lack of direct numeric optimization:} VLMs are typically trained to predict the next token rather than minimize a numeric loss function. Without tailored optimization objectives, they struggle to produce accurate and reliable regression predictions. 
\end{itemize}

To address these challenges, this paper introduces a new benchmark dataset, REO-Instruct, the first unified benchmark for regression and generation tasks in EO. We further propose REO-VLM, a novel VLM designed to unify scientific regression and generative capabilities. In summary, our contributions include:


        
    


\begin{itemize}
\item \textbf{Exploration of VLMs for multimodal EO data processing:} We conducted the first exploration of using VLMs for AGB regression from multimodal EO data, demonstrating their potential to decode complex scientific attributes and advance environmental monitoring and resource management.

\item \textbf{Creation of REO-Instruct:} We constructed REO-Instruct, the first unified dataset for regression and generation tasks in EO. It contains 1.6 million multimodal EO-language pairs, including RGB, multispectral, and SAR images, enabling joint learning of regression and generative tasks.

\item \textbf{Development of REO-VLM:} We proposed REO-VLM, a unified VLM that integrates scientific regression and generative capabilities in a single model, representing a novel approach in the EO research community.

\end{itemize}




        
    


%% file: version3/2_relatedwork.tex
\section{Related Work}
\label{sec:related work}



\subsection{VLMs for EO}
VLMs in EO can be broadly categorizing into two main streams: Contrastive VLMs and Conversational VLMs\cite{zhou2024towards}.

Contrastive VLMs primarily take both text and images as inputs and produce a similarity measure between them. This similarity is important for applications such as image-text retrieval and zero-shot scene classification. Models like~RemoteCLIP~\cite{liu2024remoteclip}, GRAFT~\cite{mall2023remote}, and SkyScript~\cite{wang2024skyscript} leverage contrastive learning mechanisms to enhance cross-modal understanding by aligning feature representations of images and text. These works primarily focus on improving modality interaction and integration.

Conversational VLMs also process text and images as inputs but output textual responses, harnessing the powerful linguistic capabilities of LLMs to support tasks like caption generation and visual question answering. These models typically comprise three main components: a pre-trained visual encoder, a pre-trained LLM, and a modality interface that connects them. The visual encoder functions similarly to the human eye, receiving and preprocessing optical signals. The LLM, akin to the human brain, understands these processed signals and performs reasoning tasks. The interface serves to align the visual and linguistic modalities. Models like RSGPT~\cite{hu2023rsgpt}, GeoChat~\cite{kuckreja2024geochat}, SkyEyeGPT~\cite{zhan2024skyeyegpt}, EarthGPT~\cite{zhang2024earthgpt}, LHRS-Bot~\cite{muhtar2024lhrs}, RS-CapRet~\cite{silva2024large}, H$^{2}$RSVLM~\cite{pang2024h2rsvlm}, RS-LLaVA~\cite{bazi2024rs}, and SkySenseGPT~\cite{luo2024skysensegpt} demonstrate the versatility and potential of conversational VLMs in diverse EO downstream tasks.

These VLM models have mainly focused on enhancing the understanding and description of image content, with limited exploration of scientific regression tasks, as shown in Table \ref{tab:method_comparison}. Nonetheless, their contributions provide valuable insights for extending VLM capabilities in EO-related research.
\begin{table}
\setlength\tabcolsep{0.6pt}
\centering
\caption{Comparison of recent methods across different tasks. LM represents large model, SM represents small model. Text means using language input or not. M-task, Reg., Cls. Cnt. denote multi-task, regression, classification and counting.}
\label{tab:method_comparison}
\begin{tabular}{lccccccc}
\toprule
Method & LM/SM & Text & M-task & Reg. & Cls. & \textbf{VQA} & \textbf{Cnt.} \\

\midrule
Niconet~\cite{lang2023high} & SM & \texttimes & \texttimes & \checkmark & \texttimes & \texttimes & \texttimes \\
Contextformer~\cite{benson2024multi} & SM & \texttimes & \texttimes & \checkmark & \texttimes & \texttimes & \texttimes \\
TorchSpatial~\cite{wu2024torchspatial} & SM & \texttimes & \checkmark & \checkmark & \checkmark & \texttimes & \checkmark \\
SRMS~\cite{hamilton2024combining} & LM & \checkmark & \texttimes & \checkmark & \texttimes & \texttimes & \texttimes \\
LHRS-bot~\cite{muhtar2024lhrs} & LM & \checkmark & \checkmark & \texttimes & \checkmark & \checkmark & \texttimes \\
GeoChat~\cite{kuckreja2024geochat} & LM & \checkmark & \checkmark & \texttimes & \checkmark & \checkmark & \texttimes \\
REO-VLM & LM & \checkmark & \checkmark & \checkmark & \checkmark & \checkmark & \checkmark \\
\bottomrule
\end{tabular}
\end{table}

\subsection{Visual Language Datasets for EO}
Training VLMs for EO requires specialized image-text datasets, which are typically developed in two main ways: creating datasets from scratch and expanding existing EO datasets.

Creating datasets from scratch involves sourcing raw EO data and pairing it with annotations. For instance, RSGPT~\cite{hu2023rsgpt} created 2,500 high-quality RSI-text pairs with manual captioning by experts. GRAFT~\cite{mall2023remote} linked ground-level image captions from social media with RSIs using geographical tags, obtaining a large dataset without manual captioning. SkyScript~\cite{wang2024skyscript} and LHRS-Align~\cite{muhtar2024lhrs} utilized OpenStreetMap to generate captions, with LHRS-Align further leveraging the language-only LLM Vicuna-v1.5 for caption production based on geographic features.

Expanding existing EO datasets involves transforming existing annotations into textual descriptions. For instance, RemoteCLIP~\cite{liu2024remoteclip} converted object detection annotations into image captions using textual templates, significantly increasing the available training data. Similarly, EarthGPT~\cite{zhang2024earthgpt} adopted a similar template-based approach. The RS5M~\cite{zhang2024rs5m} dataset, currently the largest with 5 million RSIs, employed BLIP2 to generate captions, selecting the best variants using CLIP. GeoChat~\cite{kuckreja2024geochat} constructed question-answer pairs by describing target characteristics, which were subsequently processed by a language-only LLM. Additionally, SkyEyeGPT~\cite{zhan2024skyeyegpt} combined object detection and VQA datasets to create a multitask dialogue instruction dataset.

These datasets set strong benchmarks for tasks like image captioning, VQA, and visual grounding, focusing mainly on interpreting image content. However, despite laying a solid foundation for future research, they still underutilize the potential of multimodal EO data collected from diverse sensors, particularly in addressing scientific regression challenges.

\begin{figure*}
    \centering
    \includegraphics[width=0.95\linewidth]{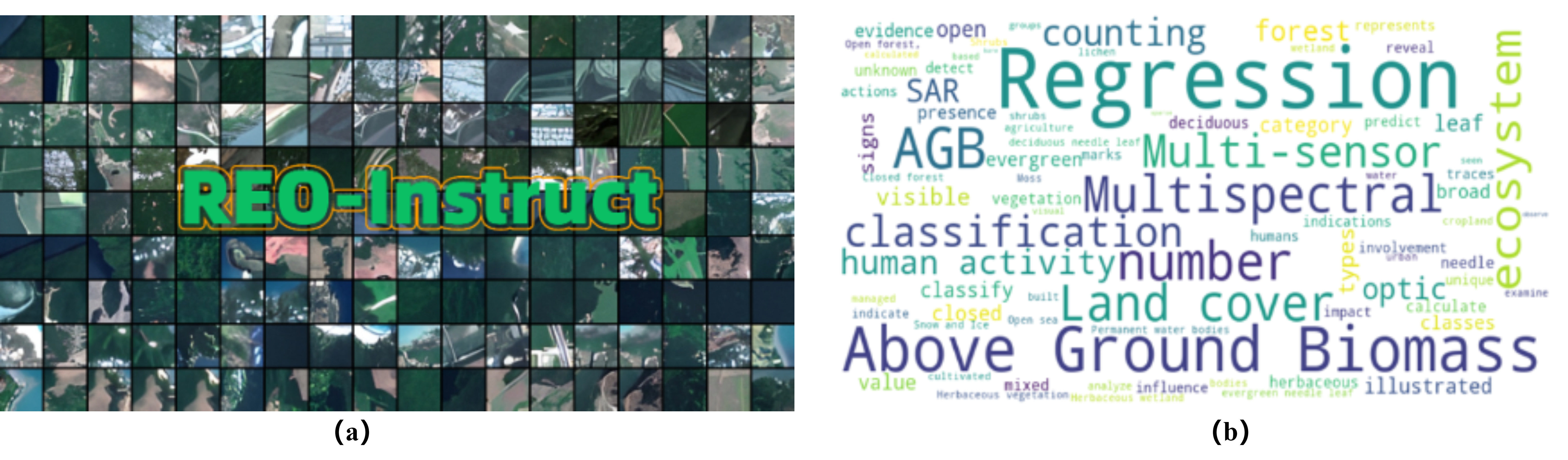}
    \caption{Image examples and prompt suite statistics of REO-Instruct benchmark. (a). Some image screenshots in RGB modality; (b). Word cloud to visualize word distribution of our prompt suites.}
    \label{fig:demo}
\end{figure*}
\subsection{Regression in EO: Emphasizing AGB Estimation}

Regression tasks play a key role in EO, enabling essential scientific applications across climatology, ecology, and geophysics. Recent advances in the computer vision community have driven significant progress in EO-related regression tasks~\cite{benson2024multi,hamilton2024combining,wu2024torchspatial}. Common tasks include species range estimation, geospatial vegetation forecasting, population density regression, forest cover prediction, nightlights intensity estimation, elevation mapping, and notably, above ground biomass (AGB) estimation~\cite{lang2023high,sialelli2024agbd}.

Several methods have emerged targeting these tasks by leveraging multimodal data and advanced learning models. For instance, Vitus et al.\cite{benson2024multi} introduced GreenEarthNet, a high-resolution vegetation forecasting dataset, along with the deep learning model Contextformer. Similarly, LE-SINR\cite{hamilton2024combining} integrates geolocations, species data, and textual descriptions into a unified framework for zero-shot species range estimation. Table~\ref{tab:method_comparison} summarizes recent methods based on model size, language integration, and task-specific capabilities in EO applications.

Given the diverse range of EO regression tasks, this paper focuses on AGB estimation as a key entry point for exploring the potential of VLMs in scientific regression. AGB~\cite{ma2021global} represents the total mass of biomass above the soil, typically measured in dry weight per unit area (e.g., grams per square meter or tons per hectare). As an essential ecological indicator, AGB plays a vital role in assessing forest carbon stocks, evaluating ecosystem health, and monitoring biodiversity.

Accurate AGB estimation is inherently challenging due to its reliance on complex environmental and anthropogenic factors. Vegetation type exerts a primary influence, as biomass density varies significantly across forests, grasslands, and agricultural lands. Additional influencing factors include forest age, human activity, climatic conditions, and soil quality. Classic AGB regression models typically rely on satellite imagery~\cite{li2020forest,laurin2018above,rodda2024lidar}, complemented by ground-truth data for model training and validation.

However, relying solely on EO imagery for AGB regression faces inherent information bottlenecks, as some key environmental and anthropogenic factors may be underrepresented. Incorporating domain-specific knowledge or intermediate interpretation results as textual inputs or embeddings could unlock new possibilities for deeper understanding and reliable prediction.

%% file: version3/3_method.tex
\section{REO-Instruct Benchmark}
\subsection{Motivation }

Developing a \textbf{unified EO-VLM capable of addressing both scientific regression tasks and image content description} holds substantial scientific and industrial significance. Benchmarks designed for EO-based regression tasks are primarily built on image-only modalities, lacking language-based annotations that could enhance models' reasoning capabilities. This limited modality design restricts the potential for incorporating domain knowledge and contextual understanding, which are crucial for complex regression tasks. Meanwhile, multimodal datasets designed for EO-VLMs always target perception, description and generation tasks, while overlooking scientific regression-oriented challenges. 

To fill this gap, we introduce the REO-Instruct benchmark, a large-scale, multimodal, and text-enriched dataset tailored for advancing EO-VLM research. It serves as a unified benchmark for model training, fine-tuning, and evaluation, enabling comprehensive assessments of both semantic understanding and scientific regression tasks in EO. By combining rich multimodal inputs with text annotations, REO-Instruct enables both perception-driven and knowledge-driven generation and regression, supporting comprehensive model learning and inference.








\subsection{Data Collection Principles}

To ensures the dataset is comprehensive and representative, we follow these three main principles:

\begin{itemize}
\item \textbf{Sufficient and necessary EO image modalities:} 
The selected EO image modalities must provide enough geospatial and spectral features, which are essential for precise forecasting into surface and vegetation characteristics. This rule ensures that the visual information is rich but not redundant, supporting accurate visual content description and AGB regression.

\item \textbf{Balanced, diverse, and representative EO image data:} The EO image data should be as balanced, diverse, and representative as possible, covering various land cover types, AGB values, human influences, and geographic distributions to build generalizable models.

\item \textbf{Scientific and domain knowledge in text annotations:} Text annotations must incorporate scientific and domain-specific knowledge relevant to description  and regression tasks, such as land cover, vegetation types, and human activity, to enhance semantic depth and prediction accuracy.
\end{itemize}





\begin{figure*}
    \centering
    \includegraphics[width=1.0\linewidth]{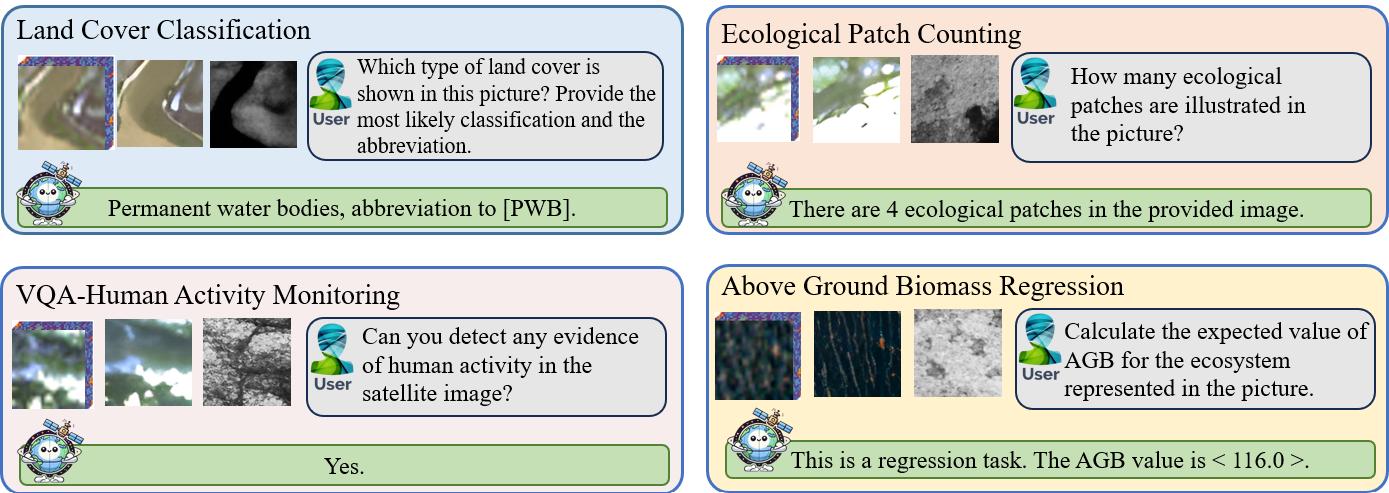}
    \caption{Screenshots of some image-texts annotation pairs in REO-Instruct benchmark. }
    \label{fig:anno}
    \vspace{-2mm}
\end{figure*}

\subsection{Overview of REO-Instruct}

REO-Instruct benchmark leverages the AGBD dataset~\cite{sialelli2024agbd}, which encompasses imagery collected during the years 2019-2020. Inspired by prior work~\cite{li2020forest,laurin2018above,rodda2024lidar},
the proposed benchmark includes three types of EO data: multispectral (MS) images, RGB three-channel optical images, and Synthetic Aperture Radar (SAR) images. The multispectral data comes from Sentinel-2 L2A, covering 13 spectral bands at a 10-meter spatial resolution. We extracted bands [4,3,2] from each multispectral image to create corresponding RGB image. The SAR data originates from ALOS-2 PALSAR-2 products, featuring two bands at a 25-meter resolution.  Images from different modalities have been spatially aligned, resulting in 25$\times$25-pixel patches corresponding to a 250m$\times$250m observation area.

These EO images captured by different sensors are further enriched with domain-specific text annotations, leveraging land cover data and the GEDI AGB data to establish text annotation. REO-Instruct benchmark comprises a significant volume of image-text pairs, with 1.6 million pairs in training set and approximately 36K pairs in testing set. Figure~\ref{fig:demo} shows image examples and prompt suite statistics of the proposed REO-Instruct benchmark.

\subsection{Text Annotations in REO-Instruct}

To generate the textual component of our dataset, we utilize ChatGPT-4o\footnote{This work invoked ChatGPT-4o, also known as GPT-4o, using OpenAI's official API.}, guided by a carefully designed prompt that ensures the inclusion of pertinent domain knowledge. As shown in Figure \ref{fig:anno}, the text annotations in the REO-Instruct benchmark cover several crucial aspects:
\begin{itemize}
    \item \textbf{Land Cover Classification:} The text annotations provide detailed descriptions of land cover types based on the Copernicus Global Land Cover Layers. Each image is assigned to one of the land cover classes, such as \textit{Closed forest, evergreen needleleaf forest}, \textit{Shrubs}, and \textit{Cultivated and Managed Vegetation/Agriculture (Cropland)}. These labels enable models to learn vegetation structures, land cover compositions, and spatial patterns critical for ecological and environmental analysis. A complete list of more than 20 land cover categories and their distributions in REO-Instruct can be found in the Supplementary Material.

     \item \textbf{Ecological Patch Counting:} The annotations provide estimates of the number of ecological patches within each observed area. An ecological patch is defined as a continuous land cover unit with distinct ecological characteristics, such as vegetation type, land use, or habitat features. This annotation reflects vegetation richness and fragmentation, where a higher number of patches indicates greater biodiversity and land cover complexity, offering critical ecological insights.
     
     \item \textbf{VQA-Human Activity Monitoring:} 
     This part of annotations include questions and answers about human-made features like urban structures, agricultural fields. These annotations discuss the potential impact of human activities on natural landscapes, facilitating the study of human-environment interactions such as urban expansion, irrigated farming, and deforestation-driven land-use change.
     \item \textbf{Above Ground Biomass Regression:} The text annotations also provide quantitative ground-truth estimates of the Above-Ground Biomass (AGB) in the corresponding image areas. These values allow models to connect textual descriptions with visual inputs, enabling direct supervision for biomass regression tasks and supporting advanced multimodal learning frameworks. Readers may refer to the supplementary for more details about AGB value distribution.
\end{itemize}

By integrating comprehensive textual annotations that include land cover properties, the number of ecological patches, human activity indicators, and direct AGB values, the REO-Instruct benchmark not only supports the development of models for AGB prediction but also enhances their capability to interpret and analyze EO data effectively. This benchmark is structured to advance the state of the art in EO analytics, leveraging both visual and textual data to foster a deeper understanding of ecological dynamics and human-environment interactions.
\subsection{Prompt Design}
To guide ChatGPT-4o in generating the most relevant answers while ensuring diversity in the responses, we have carefully designed a large number of prompts. For instance, to ensure the accuracy of land cover category descriptions, we restrict the responses to be selected from a predefined set of available categories. To ensure the diversity of questions and answers, we have set up more than 100 templates. Each conversation is generated by randomly selecting a template, while also introducing new variations to create unique responses. In addition, we categorize the questions to ensure that the regression head is invoked at the appropriate time. Meanwhile, the generation head does not produce irrelevant content, which could interfere with the regression head's decision-making. To fully utilize the data, we ensure that each image generates multiple types of question-answer dialogues. During training, one dialogue type is randomly sampled from various categories.  

\begin{figure*}
    \centering
    \includegraphics[width=1.0\linewidth]{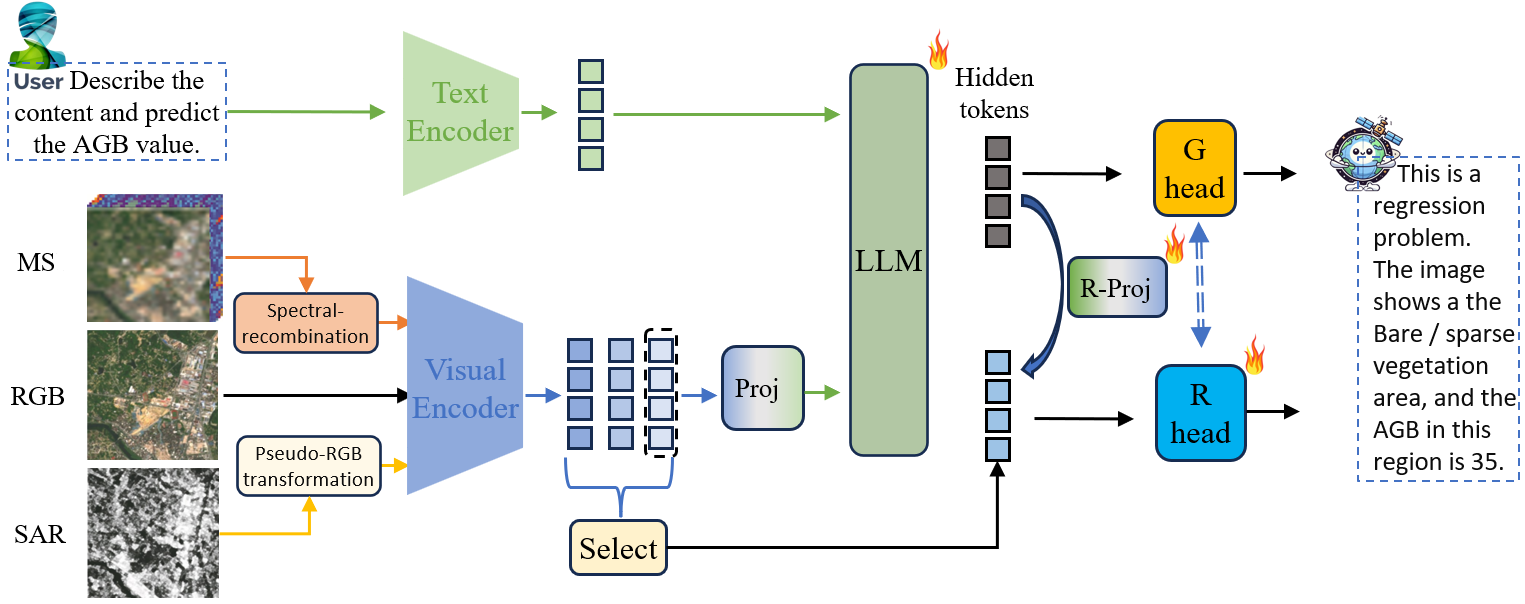}
    \caption{Overall framework of proposed REO-VLM. G head and R head denote generation and regression heads respectively. R-Proj is reverse projection module, which is in charge of pulling useful information generated by  LLM from language level to image level, jointly performing regression process. During fine-tuning, three parts marked with fire are updated. }
    \label{fig:REO-VLM}
    \vspace{-2mm}
\end{figure*}

\subsection{Challenges for VLMs in REO-Instruct}


\noindent\textbf{Heterogeneity of multimodal visual inputs:}
REO-Instruct encompasses a range of multimodal EO data, including optical and SAR imagery, each capturing distinct facets of the Earth’s surface through fundamentally different sensing mechanisms. Optical images encode reflectance values in the visible or multispectral domain, providing intuitive, camera-like views of ground features. In contrast, SAR images measure radar backscatter, yielding intensity and texture patterns shaped by object geometry, material properties, and radar illumination. Integrating such heterogeneous modalities poses significant challenges for the encoder. The model should learn to effectively fuse these disparate feature spaces, balancing the rich, visually interpretable patterns found in optical data with the often less intuitive, geometry-driven signatures of SAR imagery. Achieving this alignment demands a robust cross-modal representation strategy capable of extracting complementary information, mitigating modality-specific noise, and establishing a coherent latent space from which joint regression and generation tasks can be performed accurately.

\noindent\textbf{Misalignment between numeric regression and symbolic generation mechanisms:}
VLMs are always good at producing discrete tokens from a fixed vocabulary, excelling at semantic and contextual reasoning. In contrast, regression tasks demand mapping input features onto continuous numeric spaces with fine-grained precision. Generating a number is inherently different from producing a word, yet language models treat both as sequences of tokens from a finite set. As a result, relying on token-based generation for numeric outputs leads to instability, limited accuracy, and difficulties in achieving tight numeric control.

\noindent\textbf{Conflicting optimization objectives:} In the generative component, all representations, whether words or numbers, are treated as symbols, with the optimization goal being to generate the next most probable symbol. This aligns with the objective of generating answers in linguistic form but overlooks the inherent meaning of numbers. In contrast, the regression component focuses on the mathematical meaning of the numbers, aiming to provide accurate numerical results by fitting certain physical processes. The optimization directions for the two tasks differ, and when unified within a single framework, their optimization objectives conflict. This conflict may prevent the model from achieving optimal results. 

\noindent\textbf{Error propagation in multi-step numeric generation:}
Producing a single numeric value often involves predicting multiple tokens (e.g., generating "3.14159" character by character). Any token-level error can propagate through the sequence, compounding inaccuracies and distorting the final numeric output. Since each predicted token conditions the next, minor early inaccuracies may escalate, making high-precision numeric predictions challenging.

\section{REO-VLM Method}


Building on the REO-Instruct benchmark, we further propose REO-VLM, a unified framework that bridges the gap between visual content understanding and scientific regression tasks, enabling comprehensive EO-based analysis. Through integrating domain-specific knowledge, REO-VLM generates contextual insights, enabling deeper reasoning beyond visual features. A two-stage training strategy ensures effective multimodal feature alignment, facilitating accurate predictions and seamless task switching within a unified framework.

\subsection{Architectue}
Following pioneer works \cite{kuckreja2024geochat, zhang2024earthgpt}, we adopt LLaVA-1.5 as the foundation model. As presented in Figure \ref{fig:REO-VLM}, the proposed REO-VLM also consists of three basic components: visual encoder, text encoder, and LLM.  However, differing from LLaVA-1.5 and most of previous EO-VLM works, we optimize the basic design from the following aspects:
\begin{itemize}
    \item \textbf{Visual feature extraction:} 
    As a foundational pipeline, we aim to reuse and maximize compatibility with the pre-trained visual encoder while adapting it to EO tasks. Therefore, we introduce the spectral recombination and pseudo-RGB strategies to convert MS image and SAR image to RGB space for aligning with the pre-trained visual encoder. We split the MS into multiple three-channel images, each comprising three bands, and processed them as a group of RGB images. Given that SAR data typically contains two polarization channels, we calculate their average to create the third channel. The three resulting bands are then normalized to the [0, 255] range, generating a pseudo-RGB representation compatible with the pre-trained visual encoder. This transformation facilitates well integration without requiring additional model modifications. Although this RGB-based bridging strategy cannot fully exploit the unique characteristics of each modality, it still retains essential modality-specific information critical for downstream EO tasks.
    \item \textbf{Visual token selection:} For tasks such as generating descriptions, deep semantic information is usually sufficient to produce an accurate description. However, for regression tasks, more detailed features are often required. To achieve this, we designed a selection module that selects certain tokens from the shallow layers and adds them to the final regression process. This effectively improves regression accuracy.  The impact of different visual tokens on model performance is discussed in detail in the ablation study section.
    
    \item \textbf{Reverse projection module:} To overcome the limitations of image-only regression, we propose a reverse projection module that aligns domain-specific knowledge inferred from textual embeddings with corresponding visual features. This enables joint reasoning across modalities, breaking the information bottleneck and supporting more accurate scientific regression. 
    Inspired by the projection module in LLaVA, which adjusts image features to align with the language layer for downstream tasks, we designed a reverse projection mechanism. Specially, instead of projecting visual features into the language space, our module retrieves contextual information generated by the LLM and maps it back into the visual feature space to assist the regression process. This is achieved using a linear layer that projects 4096-dimensional hidden features from the LLM to 1024 dimensions, ensuring alignment both semantically and dimensionally.
    
    \item \textbf{Regression head:} A simple one or two layer perceptron structure is insufficient for integrating the knowledge embeddings generated by the LLM hidden tokens with the complex visual tokens. However, deeper perception architectures face optimization challenges. Therefore, inspired by \cite{tolstikhin2021mlp}, a four layer MLP-mixer like structure is designed. For specific details, please refer to the supplementary materials.

\end{itemize}

\subsection{Training strategy}
To enhance regression accuracy, we explicitly incorporate domain-specific descriptions, such as land cover types, into the text annotations for regression tasks. This design enables the generation head to learn relevant domain knowledge during training , which is then passed to the regression head through the reverse projection module, allowing for more precise and reliable predictions.

We adopt a two-stage training strategy to optimize the model's learning process. In the first stage, the LLM is trained along with the generation head. In the second stage, the focus shifts to training the regression head and reverse projection module. Such a training strategy offers several key advantages:
\begin{itemize}
\item \textbf{Conflict mitigation:}
The two-stage strategy mitigates potential conflicts stemming from differing optimization objectives between the generation and regression tasks. By decoupling these learning processes, the model maintains stability and achieves more efficient convergence.

\item \textbf{Scientific knowledge-driven inference:}
After the first stage, the generation head generates reasoning information enriched with domain-specific knowledge. During the second stage, this contextual information is integrated into the regression head through the reverse projection module, addressing the limitations of image-only regression and enhancing prediction accuracy.

\end{itemize}

Regarding the specifics of the training phases, the first stage aligns closely with the tasks handled by the pre-trained LLaVA-1.5 model, due to the similarity in the nature of the generation tasks. Therefore, we fine-tune the LLM component and generation head of the pre-trained LLaVA-1.5 model using LoRA during this phase, with cross-entropy training loss. In the second stage, which focuses on regression tasks, we load the model trained in the first stage and fine-tune the regression head and reverse projection module, using the MSE loss function as the training constraint.

As part of our pipeline, we aim to minimize the impact of non-essential factors beyond the core challenges. Hence, throughout the training process, the visual encoder and the multimodal projector remain frozen.

%% file: version3/4_experiment.tex
\section{Experimental Results}



\begin{table}
\renewcommand\arraystretch{1.0}
\setlength\tabcolsep{1.0pt}
\caption{Comparative experimental results (\%) on land cover classification task.}
\vspace{-0.2cm}
    
    \begin{threeparttable}
    \begin{tabular}{l|ccccc}
    \toprule
    Method &Modality &  OA $\uparrow$  &  MA\_Pre $\uparrow$   & MA\_Recl $\uparrow$   &  MA\_F1 $\uparrow$  \\
    \hline

    Qwen2-VL*  &RGB      & 3.77   &      0.38      &   1.14     &   0.57       \\
    ChatGPT-4o &RGB      & 3.97  &  12.27           & 3.63        &  5.60        \\
    REO-VLM &RGB      & 5.76  &  0.30           & 5.26        &  0.57        \\
    REO-VLM &MS       & 16.20 &  22.88          & 14.81       &  17.98       \\
    REO-VLM &MS+SAR  &  \textbf{19.94} &   \textbf{26.90}          &  \textbf{18.22}       &   \textbf{21.73}       \\
\bottomrule
\end{tabular}
    \begin{tablenotes}
        \item[1]\small *: metrics exclude unanswerable queries. Qwen2-VL deems 96.06\% unanswerable.
    \end{tablenotes} 
    \end{threeparttable}

\vspace{-0.3cm}
\label{tab:classification}
\end{table}

In this section, we present experimental results on the REO-Instruct benchmark to evaluate the performance of our proposed REO-VLM. We first train our model on the training set of the REO-Instruct dataset, then evaluate it against other state-of-the-art methods on distinct test subsets, each dedicated to a specific downstream task. Each downstream test subset contains approximately 8.6K unique samples, ensuring no overlap between the different subsets. The four downstream tasks include land cover classification, ecological patch counting, VQA-based human activity monitoring, and AGB regression.

For comparison, we include both domain-specific and general-purpose VLMs. Specifically, we evaluate two EO–focused large models, Geochat~\cite{kuckreja2024geochat} and LHRS-Bot~\cite{muhtar2024lhrs}, alongside general-domain VLMs including LLaVA-1.5-7B~\cite{liu2024visual}, Qwen2-VL-7B~\cite{wang2024qwen2}, and ChatGPT4o~\cite{achiam2023gpt}. In addition, we consider the results obtained when using different modalities of EO imagery as inputs to REO-VLM. This initial exploration aims to assess how varying input modalities influence inference accuracy. Unless otherwise specified, other comparison models have not been trained on our dataset. To ensure fairness during evaluation, we provided guiding prompts that explained the questions and offered a range of possible answers to other compared methods. The specific prompts are detailed in supplementary materials.


\subsection{Land cover classification results}
The comparative experimental results on land cover classification task are presented in the Table~\ref{tab:classification}. Here, OA represents overall accuracy, MA\_Pre denotes the macro-average precision score, MA\_Recl indicates the macro-average recall, and MA\_F1 stands for the macro-averaged F1 score.



\begin{figure*}
    \centering
    \includegraphics[width=0.9\linewidth]{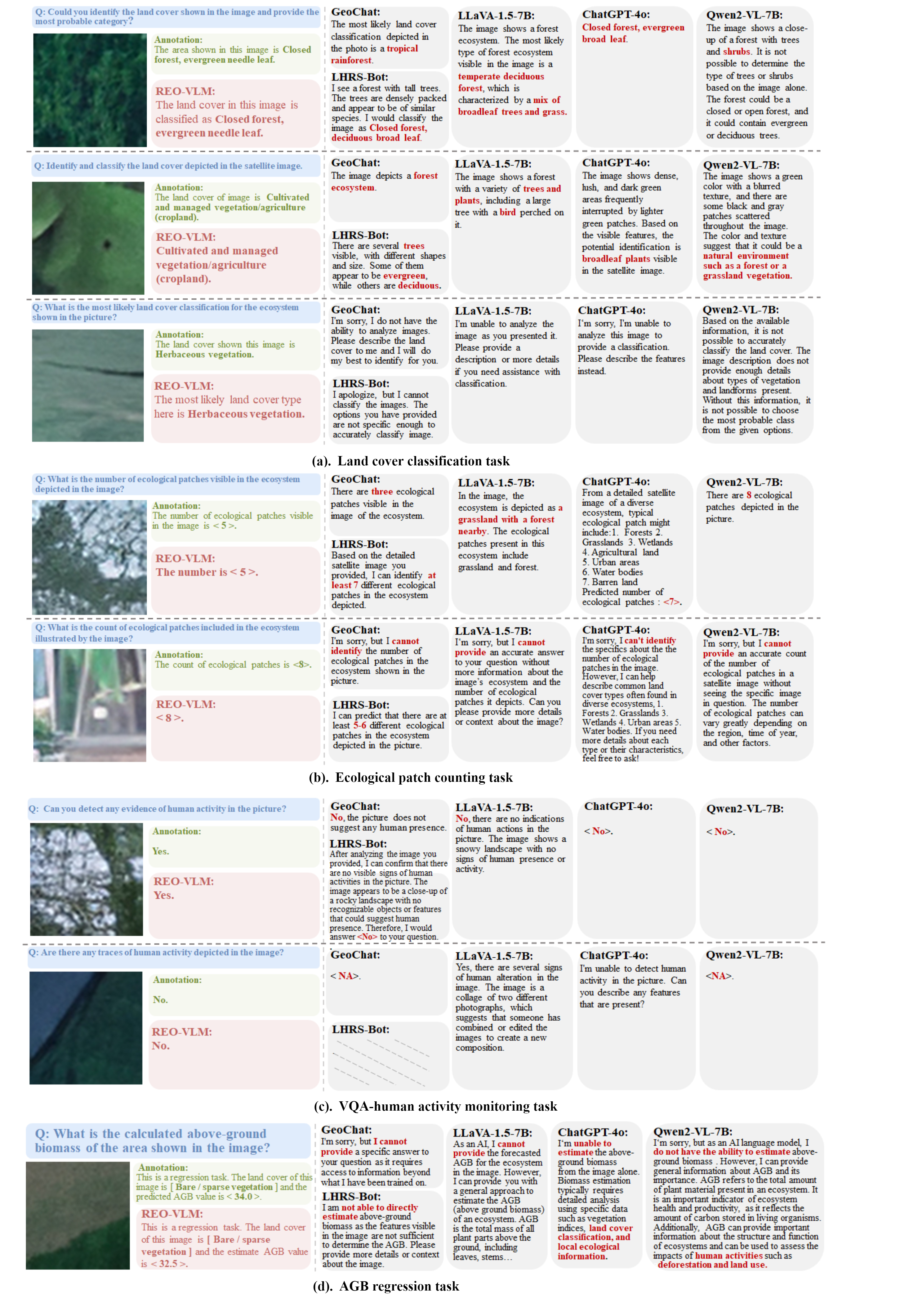}
    \vspace{-2mm}
    \caption{Qualitative experimental results of REO-VLM and other methods on different downstream tasks in REO-Instruct benchmark.}
    \label{fig:downtask_vis}
    \vspace{-2mm}
\end{figure*}

Qwen2-VL* and ChatGPT-4o, both relying solely on RGB inputs, demonstrate limited performance, with OA scores of 3.77\% and 3.97\%, respectively. Qwen2-VL* struggles significantly, attributing 96.06\% of queries as unanswerable, reflecting its limited applicability to the EO-specific task. Similarly, other comparison models, whether EO-specific (e.g., GeoChat) or general-purpose (e.g., LLaVA), also exhibit limitations in addressing land cover classification tasks, often failing to provide answers or producing unprofessional responses. Figure \ref{fig:downtask_vis}(a) presents some qualitative experimental results. We attribute these shortcomings to the lack of specific land cover domain knowledge in the training data of these models. ChatGPT-4o shows moderate improvement but still underperforms due to insufficient feature representation from RGB images alone.

In contrast, REO-VLM achieves substantial performance gains, particularly when utilizing multimodal inputs. Incorporating MS data boosts its OA to 16.20\%, indicating the value of enriched spectral information. The best results are obtained when combining MS and SAR inputs, achieving an OA of 19.94\% and a MA\_F1 of 21.73\%. This improvement highlights the model's ability to leverage complementary data sources, enhancing classification performance through more comprehensive feature extraction and robust spatial-spectral representation.

These findings demonstrate that REO-VLM's multimodal learning framework significantly enhances its capacity for land cover classification, emphasizing the critical role of multimodal data integration in EO tasks.


\begin{figure}
    \centering
    \includegraphics[width=1.0\linewidth]{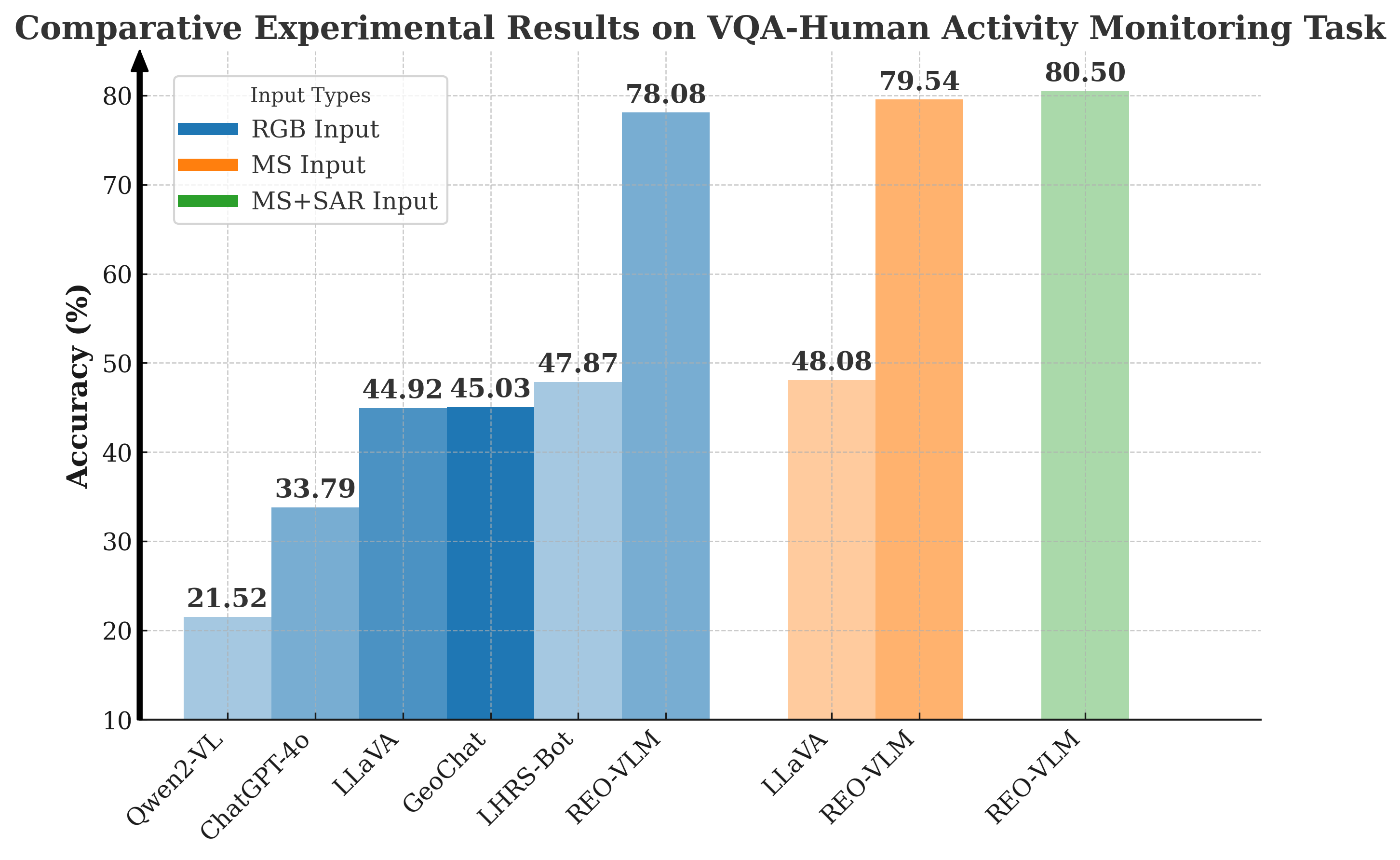}
    \caption{Comparative experimental results on the VQA-human activity monitoring task.}
    \label{fig:VQA-Human}
    \vspace{-2mm}
\end{figure}

\subsection{VQA-human activity monitoring results}
Figure \ref{fig:VQA-Human} illustrates the comparative performance of different models on the VQA-Human Activity Monitoring Task. Notably, REO-VLM outperforms both domain-specific models (GeoChat and LHRS-Bot) and general-purpose models (Qwen2-VL, ChatGPT-4o, and LLaVA) across all input modalities. Models relying solely on RGB inputs show relatively low accuracy, with Qwen2-VL achieving 21.52\% and ChatGPT-4o performing slightly better at 33.79\%. REO-VLM achieves substantial performance improvements by leveraging multimodal data. Its accuracy rises significantly from 47.87\% with RGB inputs to 79.54\% with multispectral (MS) imagery, demonstrating the value of richer spectral information. Further, integrating MS and SAR data enhances accuracy to 80.50\%, highlighting the importance of combining complementary modalities to improve inference and address complex EO tasks.


\subsection{Ecological patch counting results}
Table~\ref{tab:ecological patch} presents the comparative results on the ecological patch counting task, focusing on metrics such as RMSE, MAE, R-squared, and OA. Some qualitative analysis results are shown in Figure \ref{fig:downtask_vis}(b). While REO-VLM achieves higher predictive accuracy across all input modalities compared to other models, the negative R-squared values across the board reveal a critical limitation. This results highlights that none of the models, including REO-VLM, successfully captured the underlying patterns of the data. 

The root cause lies in the approach used for numeric prediction. In this experiment, ecological patch counts were treated as texts, and the models were trained using cross-entropy loss, a optimization method better suited for discrete classification than continuous numeric regression. 

This result provides an important insight: treating numeric regression as token-by-token text generation is fundamentally flawed. The cross-entropy optimization mechanism struggles to align with the requirements of accurate numeric prediction, further emphasizing the need for tailored approaches, such as regression-specific loss functions or numeric-aware architectures, for effectively handling numeric tasks.


\begin{table}
\renewcommand\arraystretch{1.0}
\setlength\tabcolsep{1.2pt}
\caption{Comparative experimental results on ecological patch counting  task.}
\vspace{-0.2cm}
    
    \begin{threeparttable}
    \begin{tabular}{l|ccccc}
    \toprule
    Method &Modality   & RMSE$\downarrow$   & MAE$\downarrow$ &  R-squared$\uparrow$   & OA (\%)$\uparrow$ \\
    \hline
    Qwen2-VL* &RGB   & 12.74 & 4.30  & -121.79    &9.75   \\
    ChatGPT-4o &RGB      &  5.13  & 4.79 &  -18.42      & 2.43    \\
    LLaVA*\ &RGB           &  1.31    & 1.06     & -0.27  &  25.34       \\
    REO-VLM &RGB      &   \textbf{1.23}  &  \textbf{0.91} &   \textbf{-0.12}      & 32.47       \\
    
    LLaVA*  &MS        &  1.38    &  1.13    &  -0.40 &  22.67  \\
    REO-VLM &MS       &  1.39  & 0.99 &  -0.43      & 34.79       \\
    REO-VLM &MS+SAR   &  1.27  &  \textbf{0.91} &  -0.19      &  \textbf{35.90}     \\
\bottomrule
\end{tabular}
    \begin{tablenotes}
        \item[1]\small
*: metrics exclude unanswerable queries. LLaVA answers 79.31\% of RGB and 21.20\% of MS queries, while Qwen2-VL only answers 20.31\% questions.   %
    \end{tablenotes} 
    \end{threeparttable}

\vspace{-0.3cm}
\label{tab:ecological patch}
\end{table}

\begin{table}
\renewcommand\arraystretch{1.0}
\setlength\tabcolsep{5.4pt}
\caption{Comparative experimental results on AGB regression task.}
\vspace{-0.2cm}
    \label{table2}
    \begin{threeparttable}
    \begin{tabular}{l|ccccc}
    \toprule
    Method &Modality   & RMSE$\downarrow$    & MAE$\downarrow$  &  R-squared$\uparrow$ \\
    \hline
    LLaVA\dag \ &RGB      & 116.60  &   67.90         &    -0.51             \\
    REO-VLM &RGB       &  85.10  & 50.56 &  0.20       \\
    LLaVA*\dag  &MS      &  115.74 &  67.49          &   -0.45               \\
    REO-VLM &MS       &  \textbf{75.59}  & \textbf{48.37} &  \textbf{0.36}        \\
    REO-VLM &MS+SAR   &  76.52  & 43.84 &  0.35      \\
\bottomrule
\end{tabular}
    \begin{tablenotes}
        \item[1]\small \dag: Models fine-tuned on REO-Instruct.\\
*: Unanswerable cases excluded;  LLaVA*\dag only counted 89.31\% of questions with definite answers.
    \end{tablenotes} 
    \end{threeparttable}
\vspace{-0.3cm}
\label{tab:agb}
\end{table}

\subsection{AGB Regression results}

Table~\ref{tab:agb} presents the comparative experimental results for the AGB regression task. As shown in Figure \ref{fig:downtask_vis}(d), almost all comparison algorithms fail to perform effective AGB regression. To validate the effectiveness of the REO-VLM architecture and optimization strategy, we fine-tuned the original LLaVA model on the REO-Instruct dataset. However, the fine-tuned LLaVA model still exhibits a negative R-squared value across both RGB and MS modalities, indicating that it failed to learn meaningful numeric patterns from the EO data and domain knowledge.

In contrast, the proposed REO-VLM achieves significantly better performance. By decoupling regression and generative task heads and employing a two-stage training strategy, REO-VLM demonstrates improved performance across all modalities. Specifically, REO-VLM achieves a positive R-squared value (0.20 for RGB, 0.36 for MS, and 0.35 for MS+SAR), proving that the model captures underlying numerical patterns in the data. Additionally, REO-VLM achieves lower RMSE and MAE compared to fine-tuned LLaVA, further highlighting its superior regression capabilities. These results underscore the effectiveness of the proposed architectural and optimization enhancements in addressing the challenges of AGB regression.

From the experimental results, it is clear that AGB prediction has a strong dependence on multispectral data. When using multispectral data as input, our model achieved an RMSE of 75.59, which indicates that using VLM to predict AGB is feasible. 


In fact, building a single-task model for AGB regression prediction is highly challenging. The best model in this field is Niconet \cite{lang2023high}, which achieved 69.0 on RSME metric on REO-Instruct dataset. Compared with Niconet, our REO-VLM shows a slight performance gap in AGB prediction, while covering a broader range of tasks.

\subsection{Ablation study}

In Table \ref{tab:ablation}, we test the impact of extracting visual feature tokens from different layers on the AGB regression task. There are four strategies: 1) Extract features only from the last layer; 2) Extract features from every two layers; 3) Continue extracting features from half of the layers but with a sparse extraction from shallow layers and denser extraction from deeper layers; 4) Input all visual layers into the LLM.

From the experimental results, it is clear that using multi-layer features in a balanced manner yields the best performance. This experimental result also supports our previous analysis, indicating that tasks like AGB prediction are highly sensitive to detailed information. Compared to using only the features from the last visual layer, giving equal weight to both deep and shallow features resulted in a significant performance improvement. However, when all the visual features are extracted and fed into the LLM for the final regression prediction, the overall performance drops significantly. We speculate that this is due to the excessive amount of visual information, which disrupts the model's optimization and training process. This not only makes it difficult to generate useful auxiliary information but also further affects the convergence of the generation head. 


\begin{table}
\setlength\tabcolsep{1pt}
\renewcommand\arraystretch{1.0}
\centering
\caption{Experimental result with different visual token selection strategies}
\begin{tabular}{l|ccccc}
\toprule
Token layer   & Modality    &  RMSE$\downarrow$  &  MAE$\downarrow$   &  R-squared$\uparrow$ \\
\hline
Last layer                &  MS&  87.21 &  52.67 &  0.16      \\
Half layers                  &  MS  &  \textbf{75.59} &  \textbf{48.37} &  \textbf{0.36}      \\
Half layers (deep)     &  MS &  77.55 &  48.50 & 0.33   \\
All layer            &  MS & 117.19 & 68.71 &  -0.52     \\
\bottomrule
\end{tabular}
\label{tab:ablation}
\end{table}

%% file: version3/5_conclusion.tex
\section{Conclusion}
In this work, we revisit the current state of VLM applications in EO. 
We analyzed some of the existing overlooked aspects of VLM in EO applications, particularly its limited exploration in regression tasks. One key challenge is the lack of mature datasets that support learning domain-specific knowledge for regression. Additionally, new mechanisms are needed to improve compatibility with complex regression tasks in EO scenarios. In corresponding to these challenges, we firstly construct a large scale benchmark, named REO-Instruct. It covers 1.6 million multimodal visual—language pairs, laying the foundation for future work. Then we propose REO-VLM, the first unified EO-VLM model, which can handle both the challenges of regression and traditional generation tasks. 


\noindent\textbf{Limitations and Future work.}
Despite these advancements, this work still faces constraints. The relatively small image sizes limit the level of spatial detail that can be captured, potentially restricting the model’s ability to fully represent complex environmental patterns. Building on this foundation, future research could integrate higher-resolution imagery and incorporate additional data sources, such as hyperspectral, LiDAR, or time-series observations to improve detail and robustness. Exploring uncertainty quantification and explainability methods presents a promising research direction, offering potential for more reliable and interpretable predictions. Additionally, investigating multi-step scientific regression strategies could enable EO-VLMs to tackle more complex scientific scenarios, opening up promising opportunities for addressing intricate environmental and ecological challenges.

%% file: version3/appendix.tex
\newpage
\maketitlesupplementary
\setcounter{section}{0}
\begin{figure}
    \centering
    \includegraphics[width=1.0\linewidth]{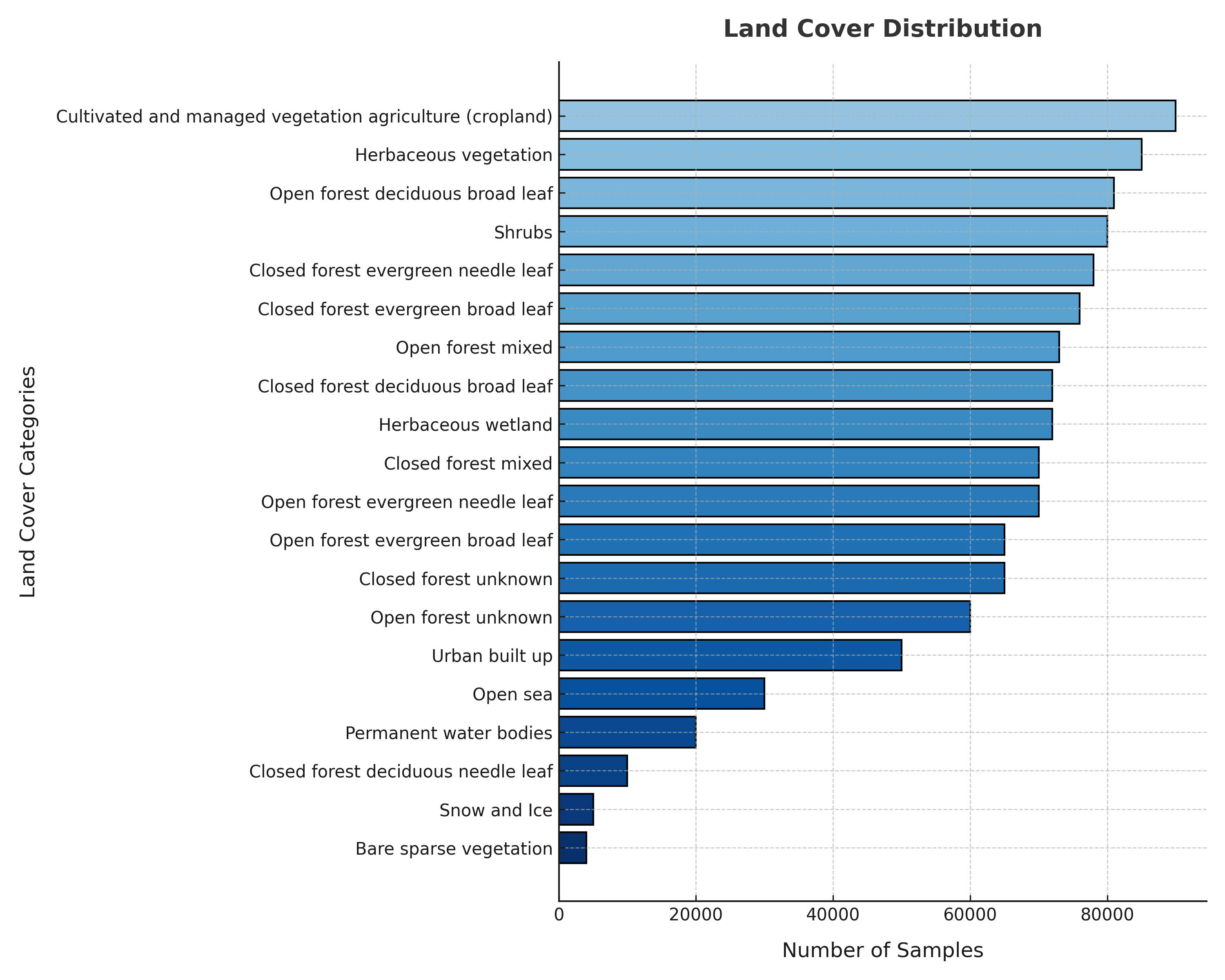}
    \caption{Land cover distribution based on the number of samples in each category. }
    \label{fig:Lc-distribution}
    \vspace{-2mm}
\end{figure}
\vspace{-10mm}
\section{Land Cover Distribution in REO-Instruct }
Figure~\ref{fig:Lc-distribution} illustrates the distribution of land cover categories based on the number of samples collected for each type. The categories, ranging from "Cultivated and managed vegetation" to "Bare sparse vegetation," are sorted in descending order. This visualization highlights the significant representation of agricultural and vegetative covers, such as cropland and herbaceous vegetation, while categories like "Snow and Ice" and "Bare sparse vegetation" have relatively fewer samples. Compared with the land cover distribution in the original AGBD~\cite{sialelli2024agbd} dataset, we ensure the balance between different land cover categories as much as possible by weighted selection and resampling.
\vspace{-2mm}
\section{Above Ground Biomass Value Distribution}
\begin{figure}
    \centering
    \includegraphics[width=1.0\linewidth]{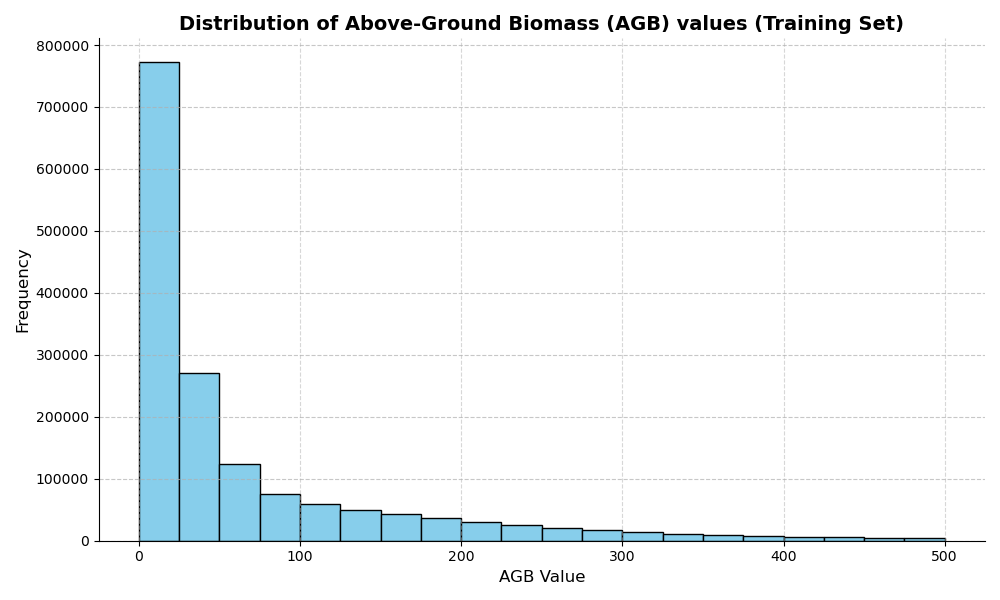}
    \includegraphics[width=1.0\linewidth]{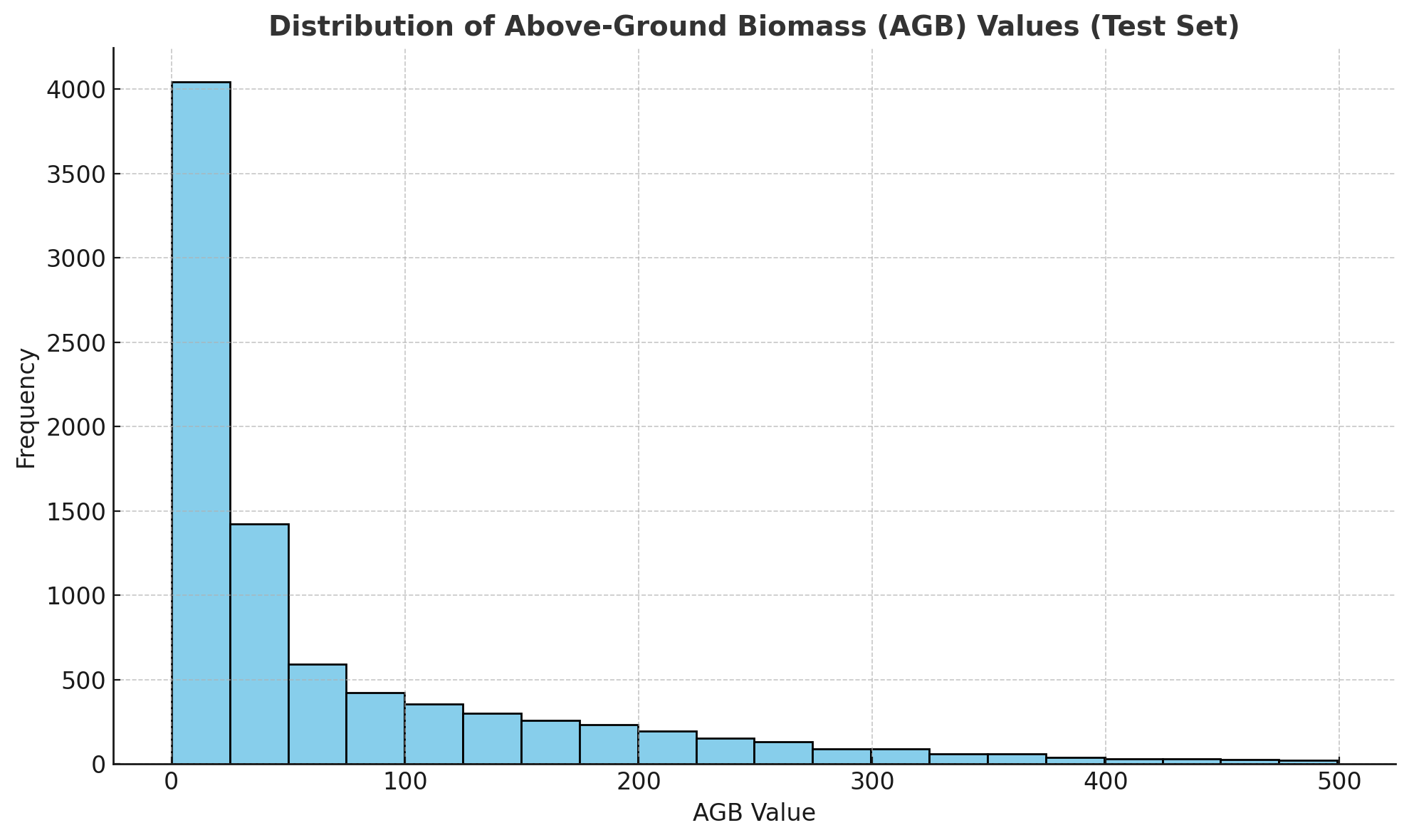}
    \caption{Distribution of Above-Ground Biomass (AGB) values in REO-Instruct. The histogram shows the frequency of AGB values.}
    \label{fig:agb-distribution}
    \vspace{-2mm}
\end{figure}

Figure \ref{fig:agb-distribution} shows the distribution of AGB values in both the training and test sets. The data distributions in the two sets are highly similar, ensuring consistency and fairness between the two sets. The AGB values range from [0-500], following a long-tail distribution where larger values correspond to fewer samples. This imbalance further increases the difficulty of  complex scientific regression task.
\begin{figure*}
    \centering
    \includegraphics[width=.95\linewidth]{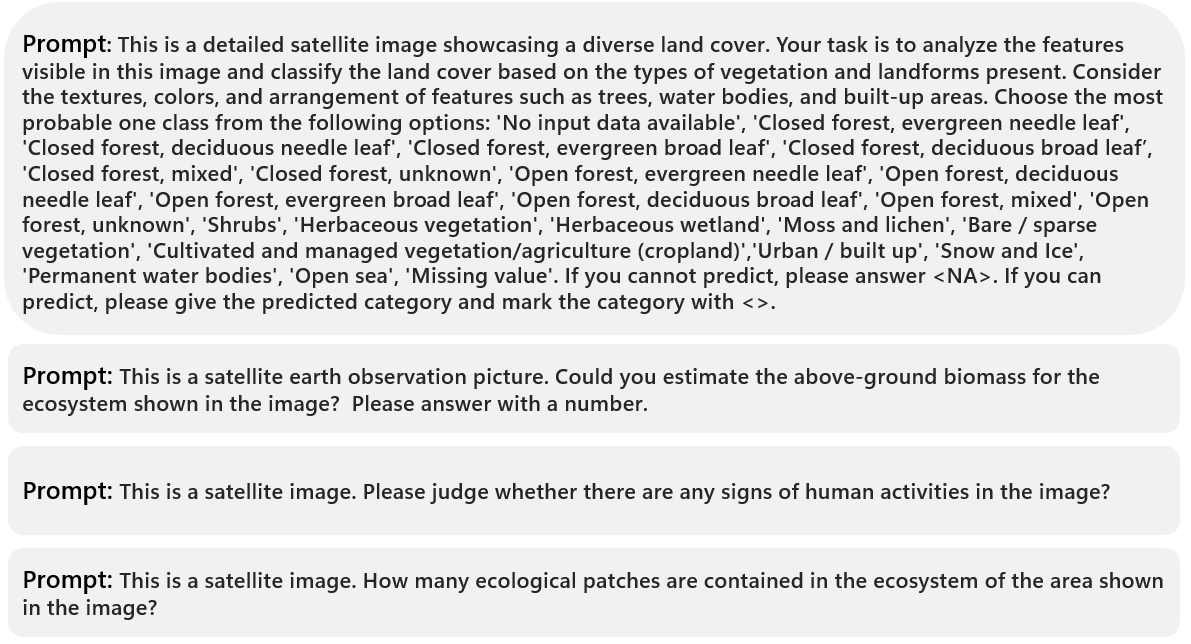}
    \caption{Test guiding prompts for compared methods.}
    \label{fig:prompts}
    \vspace{-2mm}
\end{figure*}

\section{Spectral Recombination Strategy}
In the field of EO, band combinations are commonly used to better interpret image features by enhancing specific characteristics. By selecting appropriate band combinations, we can extract relevant information from images, such as highlighting geological, agricultural, or vegetation features. Inspired by this, we reorganize each MS image into five distinct three-channel images using the following band recombination strategies:
\begin{itemize}
    \item \textbf{Red-edge Bands [B05, B06, B07]:} These bands are highly effective for monitoring vegetation health by capturing subtle changes in plant reflectance.
    \item \textbf{Geology Bands [B12, B11, B02]:}
This combination highlights geological features such as faults, lithology, and geological structures, making it valuable for geological mapping.
    \item \textbf{Natural Color Bands [B04, B03, B02]:}
These bands capture many visually interpretable features, producing natural-looking RGB images similar to human vision.
    \item \textbf{Color Infrared Bands [B08, B04, B03]:}
This combination distinguishes between healthy and unhealthy vegetation, which reflects chlorophyll strongly. In standard false-color images, dense vegetation appears red, while urban areas are white.
    \item \textbf{Short-Wave Infrared Bands [B12, B08, B04]:}
This combination highlights various shades of green representing vegetation density. Darker green shades indicate denser vegetation, while brown shades suggest exposed soil or built-up areas.
\end{itemize}

By leveraging these recombination strategies, we enhance feature representation while enabling the reuse of pretrained visual encoders. Compared to randomly assembling pseudo-RGB images from different bands of MS images, our approach facilitates more effective interpretation and analysis in EO tasks.





\section{Regression Head Architecture}

The visual tokens from the encoder and the hidden tokens from the LLM are combined and processed through the regression head. Every layer in the regression head includes two main stages:

\subsection{Token Aggregation Stage}
This stage fuses intra-token features by combining linear transformations and normalization. For each token, the operation is expressed as:
\begin{equation}
u_i = \sigma \big( w_1 \cdot \mathcal{L}(x) \big) + w_2 \cdot x,
\end{equation}
where \(x\) represents the input visual tokens, \(\sigma\) is an element-wise nonlinearity, and \(w_1, w_2\) are learnable parameters. Here, \( \mathcal{L}(x) \) compactly represents the layer normalization function, which normalizes the input features for improved stability and performance.

\subsection{Knowledge Integration Stage}
The aggregated features \(u\) are refined to integrate domain knowledge and enable interaction between different tokens, facilitating cross-modal and intra-token information exchange. This operation is defined as:
\begin{equation}
y_j = \sigma \big( w_3 \cdot \mathcal{L}(u) \big) + w_4 \cdot u,
\end{equation}
where \(w_3, w_4\) are learnable weights applied to the intermediate features.

After processing, the final output tokens are averaged to produce regression result. This design mitigates overfitting. 

Unlike models designed for classification task, Dropout operation is omitted here as it introduces instability in numerical regression tasks. 
Specifically, if dropout is applied, the variance of hidden neuron outputs during training becomes inconsistent with that during test. This variance shift, after passing through the non-linear mapping layer, causes output value bias, ultimately leading to poor performance on the test set. Experiments also demonstrate that this design ensures better stability and performance.

\section{Guiding Prompts}
To define the scope of our questions and expected answers, we provide guiding prompts during testing for all comparison algorithms, except for our proposed method. These prompts shown in Figure \ref{fig:prompts} assist VLMs in better understanding and addressing multiple tasks.